\documentclass[runningheads]{llncs}
\usepackage{graphicx}
\usepackage{dirtytalk}
\usepackage{amsmath,amssymb,amsfonts}
\usepackage{algorithmic}
\usepackage{algorithm2e}

\DeclareMathOperator*{\argmax}{arg\,max}

\begin{document}
\title{Pseudo-Label Selection is a Decision Problem\thanks{Extended abstract accepted for presentation at the \textit{46th German Conference on Artificial Intelligence}, Berlin, Germany, September 26–29, 2023. }}
%
%
\author{Julian Rodemann\thanks{Julian Rodemann thanks his collaborators Thomas Augustin, Jann Goschenhofer, Emilio Dorigatti, Thomas Nagler, Christoph Jansen and Georg Schollmeyer. Furthermore, he gratefully acknowledges support by the Federal Statistical Office of Germany within the co-operation project 'Machine Learning in Official Statistics', the LMU mentoring programm, and the bidt graduate center by the Bavarian Academy of Sciences (BAS).} 
}
\authorrunning{J. Rodemann}
%
\institute{    Department of Statistics\\
    Ludwig-Maximilians-Universität (LMU)\\
    Munich, Germany \\
\email{julian@stat.uni-muenchen.de}\\
}
\maketitle              
\begin{abstract}

Pseudo-Labeling is a simple and effective approach to semi-supervised learning. It requires criteria that guide the selection of pseudo-labeled data. The latter have been shown to crucially affect pseudo-labeling's generalization performance. Several such criteria exist and were proven to work reasonably well in practice. However, their performance often depends on the initial model fit on labeled data. Early overfitting can be propagated to the final model by choosing instances with overconfident but wrong predictions, often called confirmation bias. 

In two recent works, we demonstrate that pseudo-label selection (PLS) can be naturally embedded into decision theory. This paves the way for BPLS, a Bayesian framework for PLS that mitigates the issue of confirmation bias \cite{rodemann2023-bpls}. At its heart is a novel selection criterion: an analytical approximation of the posterior predictive of pseudo-samples and labeled data. We derive this selection criterion by proving Bayes-optimality of this \say{pseudo posterior predictive}. We empirically assess BPLS for generalized linear, non-parametric generalized additive models and Bayesian neural networks \mbox{on} simulated and real-world data. When faced with data prone to overfitting and thus a high chance of confirmation bias, BPLS outperforms traditional PLS methods. 
The decision-theoretic embedding further allows us to render PLS more robust towards the involved modeling assumptions \cite{inalllikelihoods}. To achieve this goal, we introduce a multi-objective utility function. We demonstrate that the latter can be constructed to account for different sources of uncertainty and explore three examples: model selection, accumulation of errors and covariate shift.


 \\
 
 \end{abstract}

\section{Introduction: Pseudo-Labeling}

Labeled data are hard to obtain in a good deal of applied classification settings. This has given rise to the paradigm of semi-supervised learning (SSL), where information from unlabeled data is (partly) taken into account to improve inference drawn from labeled data in a supervised learning framework. 
Within SSL, an intuitive and widely used approach is referred to as self-training or pseudo-labeling \cite{shi2018transductive,lee2013pseudo,rizve2020defense}. The idea is to fit an initial model to labeled data and iteratively assign pseudo-labels to some of the unlabeled data according to the model's predictions. This process requires a criterion for pseudo-label selection (PLS)\footnote{Other names: Self-Training, Self-Labeling.}, that is, the selection of pseudo-labeled instances to be added to the training data.

\RestyleAlgo{ruled}
\SetKwComment{Comment}{/* }{ */}

\begin{algorithm}[H]
\caption{Pseudo-Labeling (Self-Training, Self-Labeling)}
\label{alg:main}

\KwData{$\mathcal{D}, \mathcal{U}$}
\KwResult{fitted model $\hat y^*(x)$}

\While{stopping criterion not met}{
\textbf{fit} model on labeled data $\mathcal{D}$ to obtain prediction function $\hat y(x)$ \\
\For{$i \in \{1, \dots, \lvert \mathcal{U} \rvert \}$}{
\textbf{predict} $\mathcal{Y} \ni \hat y_i = \hat y(x_i)$ with $x_i$ from $\left(x_{i}, \mathcal{Y}\right)_i$ in $\mathcal{U}$ \\
\textbf{compute} some selection criterion $c(x_{i}, \hat y_i)$
\\
}
\textbf{obtain} $i^* = \argmax_i \, c(x_{i}, \hat y_i) \, $ \\ 
\textbf{add} $(x_{i^*}, \hat y_{i^*})$ to labeled data: $\mathcal{D} \leftarrow \mathcal{D} \cup (x_i, \hat y_i) $ \\
\textbf{update} $\mathcal{U} \leftarrow \mathcal{U} \setminus \left(x_{i}, \mathcal{Y}\right)_i $
}
\end{algorithm}

Various selection criteria have been proposed in the literature \cite{shi2018transductive,lee2013pseudo,rizve2020defense}. However, there is hardly any theoretical justification for them. By embedding PLS into decision theory, we address this research gap. Another issue with PLS is the reliance on the initial model fit on labeled data. If the initial model generalizes poorly, misconceptions can propagate throughout the process \cite{arazo2020pseudo}. Accordingly, we exploit Bayesian decision theory to derive a PLS criterion that is robust with respect to the initial model fit, thus mitigating this so-called confirmation bias.

\section{Approximatley Bayes-optimal Pseudo-Label Selection}

We argue that PLS is nothing but a canonical decision problem: We formalize the selection of data points to be pseudo-labeled as decision problem, where the unknown set of states of nature is the learner's parameter space and the action space -- unlike in statistical decision theorey -- corresponds to the set of unlabeled data, i.e., we regard (the selection of) data points as actions.
This perspective clears the way for deploying several decision-theoretic approaches -- first and foremost, finding Bayes-optimal actions (selections of pseudo-labels) under common loss/utility functions. In our first contribution \cite{rodemann2023-bpls}, we prove that with the joint likelihood as utility, the Bayes-optimal criterion is the posterior predictive of pseudo-samples and labeled data. Since the latter requires computing a possibly intractable integral, we come up with approximations based on Laplace's method and the Gaussian integral that circumvent expensive sampling-based evaluations of the posterior predictive. Our approximate version of the Bayes-optimal criterion turns out to be simple and computationally cheap to evaluate: $ \ell (\hat \theta) - \frac{1}{2} \log \lvert \mathcal{I}(\hat \theta) \rvert $ with $\ell(\hat \theta)$ being the likelihood and $\mathcal{I}(\hat \theta)$ the Fisher-information matrix at the fitted parameter vector $\hat \theta$. As an approximation of the joint posterior predictive of pseudo-samples and labeled data, it does remarkably not require an \textit{i.i.d.} assumption. This renders our criterion applicable to a great variety of applied learning tasks. 

We deploy BPLS on simulated and real-world data using several models ranging from generalized linear over non-parametric generalized additive models to Bayesian neural networks. Empirical evidence suggests that such a Bayesian approach to PLS -- which we simply dub Bayesian PLS (BPLS) -- can mitigate the confirmation bias \cite{arazo2020pseudo} in pseudo-labeling that results from overfitting initial models. Besides, BPLS is flexible enough to incorporate prior knowledge not only in predicting but also in selecting pseudo-labeled data. What is more, BPLS involves no hyperparameters that require tuning. Notably, the decision-theoretic treatment of PLS also yields the framework of optimistic (pessimistic) superset learning \cite{hullermeier2014learning,hullermeier2019learning,rodemann2022supersetlearning} as max-max- (min-max-)actions.


\section{In All Likelihood\textit{s}: Robust Pseudo-Label Selection}

The decision-theoretic embedding opens up a myriad of possible venues for future work thanks to the rich literature on decision theory. In our second contribution \cite{inalllikelihoods}, we propose three extensions of BPLS by leveraging results from generalized Bayesian decision theory. Our extensions aim at robustifying PLS with regard to model selection, accumulation of errors and covariate shift \cite{rodemann2022not}. The general idea is to define a multi-objective utility function. The latter can consist of likelihood functions from, e.g., differently specified models, thus incorporating potentially competing goals. We further discuss how to embed such multi-objective utilities into a \textit{preference system} \cite{jsa2018,uai2023_all}. This allows us to harness the entire information encoded in its cardinal dimensions while still being able to avoid unjustified assumptions on the hierarchy of the involved objectives. 
We further consider the generic approach of the generalized Bayesian $\alpha$-cut updating rule for credal sets of priors. Such priors can not only reflect uncertainty regarding prior information, but might as well represent priors near ignorance, see~\cite{benavoli2015prior,mangili2015new,mangili15prior,rodemann-BO-isipta,prior-robust-BO} for instance. 

We spotlight the application of the introduced extensions on real-world and simulated data using semi-supervised logistic regression. In a benchmarking study, we compare these extensions to traditional PLS methods. Results suggest that especially robustness with regard to model choice can lead to substantial performance gains.

%
%
 \bibliographystyle{splncs04}
%
\bibliography{bib}

\end{document}